\documentclass[nohyperref]{article}

\usepackage{stackrel}
\usepackage{microtype}
\usepackage{graphicx}
\usepackage{subfigure}
\usepackage{booktabs} % for professional tables
\usepackage{comment}
\usepackage{hyperref}

\usepackage[accepted]{icml2023}

\usepackage{amsmath}
\usepackage{amssymb}
\usepackage{mathtools}
\usepackage{amsthm}
\usepackage{bm}
\usepackage[normalem]{ulem}

\newcommand{\bx}[0]{{\bm{x}}}

\newcommand{\bt}[0]{{\bm{\theta}}}
\newcommand{\bJ}[0]{\bm{J}}
\newcommand{\bh}[0]{\bm{h}}

\newcommand{\der}[0]{{\mathrm{d}}}
\newcommand{\U}[0]{{\mathcal{U}}}
\newcommand{\bU}[0]{\bm{\mathcal{U}}}
\newcommand{\bu}[0]{\bm{u}}
\newcommand{\like}[0]{{\mathcal{L}}}
\newcommand{\pemp}[0]{{p_{\mathcal{D}}}}
\newcommand{\caja}[1]{\left[ {#1} \right] }
\newcommand{\paren}[1]{\left( {#1} \right) }

\newcommand{\av}[1]{\left\langle {#1} \right\rangle }

\usepackage{xcolor}
\definecolor{darkblue}{rgb}{0,0,0.6}
\definecolor{darkred}{rgb}{0.6,0,0}

\definecolor{myblue}{RGB}{12, 12, 158}
\definecolor{myred}{RGB}{158, 19, 22}
\definecolor{myorange}{RGB}{245, 150, 12}
\definecolor{mygreen}{RGB}{26, 148, 49}
\definecolor{Prune}{RGB}{99,0,60}
\definecolor{Purple}{RGB}{75, 0, 130}
\definecolor{Pink}{RGB}{255, 105, 180}
\definecolor{deepskyblue}{RGB}{0, 191,255}
\definecolor{limegreen}{RGB}{50, 205, 50}
\definecolor{crimson}{rgb}{0.86, 0.08, 0.24}
\definecolor{coral}{rgb}{1.0, 0.5, 0.31}

\definecolor{blue(ncs)}{rgb}{0.0, 0.53, 0.74}

\usepackage{hyperref}
\hypersetup{
bookmarksopen=true,
pdftitle="",
pdfauthor="", %whatever names should be relevant
pdfsubject="", %subject of the document
pdfstartview={FitH},		% fits the width of the page to the window
pdfmenubar=true,			%menubar shown
pdfhighlight=/O,			%effect of clicking on a link
colorlinks=true,			%couleurs sur les liens hypertextes
urlcolor=darkblue,
citecolor=darkblue,		%couleur des liens pour les citations
linkcolor=Prune,	%couleur des liens hypertextes internes
}

\usepackage[capitalize,noabbrev]{cleveref}

\newtheorem{theorem}{Theorem}[section]

\theoremstyle{definition}

\theoremstyle{remark}

\usepackage[textsize=tiny]{todonotes}

\icmltitlerunning{~ \hfill Explaining the effects of non-convergent sampling in the training of Energy-Based Models \hfill \thepage}

\begin{document}

\twocolumn[
%\icmltitle{Learning with non-convergent Monte-Carlo dynamics}
%\icmltitle{Explaining non-convergent Monte Carlo sampling effects in maximum likelihood learning of energy-based models}
%\icmltitle{Sampling issues in maximum likelihood learning of energy-based models}
%\icmltitle{Non-convergent MCMC Sampling effects in the learning of Energy-Based Models}

%\icmltitle{Impact of short MCMC runs in the training of Energy-Based Models}
\icmltitle{Explaining the effects of non-convergent MCMC in the training of Energy-Based Models}
%\icmltitle{Generating perfect samples using non-convergent MCMC in the training of Energy-Based Models}



\icmlsetsymbol{equal}{*}

\begin{icmlauthorlist}
\icmlauthor{Elisabeth Agoritsas}{geneva}
\icmlauthor{Giovanni Catania}{mad}
\icmlauthor{Aurélien Decelle}{mad,paris}
\icmlauthor{Beatriz Seoane}{mad,paris}
\end{icmlauthorlist}

\icmlaffiliation{geneva}{Department of Quantum Matter Physics, University of Geneva, 1211 Geneva, Switzerland}
\icmlaffiliation{mad}{Departamento de Física Teórica, Universidad Complutense de Madrid, 28040
Madrid, Spain}
\icmlaffiliation{paris}{Université Paris-Saclay, CNRS, INRIA Tau team, LISN, 91190 Gif-sur-Yvette,
France}

\icmlcorrespondingauthor{Aurélien Decelle}{adecelle@ucm.es}
\icmlcorrespondingauthor{Beatriz Seoane}{beseoane@ucm.es}

\icmlkeywords{Machine Learning, ICML}

\vskip 0.3in
]



\printAffiliationsAndNotice{\icmlEqualContribution} % otherwise use the standard text.

\begin{abstract}
In this paper, we quantify the impact of using non-convergent Markov chains to train Energy-Based models (EBMs). In particular, we show analytically that EBMs trained with non-persistent short runs to estimate the gradient can perfectly reproduce a set of empirical statistics of the data, not at the level of the equilibrium measure, but through a precise dynamical process. Our results provide a first-principles explanation for the observations of recent works proposing the strategy of using short runs starting from random initial conditions as an efficient way to generate high-quality samples in EBMs, and lay the groundwork for using EBMs as diffusion models. After explaining this effect in generic EBMs, we analyze two solvable models in which the effect of the non-convergent sampling in the trained parameters can be described in detail. Finally, we test these predictions numerically on a ConvNet EBM and a Boltzmann machine.\end{abstract}

\section{Introduction}

Encoding the statistics of a given dataset and generating data points that match those statistics are crucial issues in machine learning.
Many different strategies  have been devised. %proposed.
Among the most popular solutions are Variational Autoencoders (VAEs)~\cite{kingma2018glow}, Generative Adversarial Networks (GANs)~\cite{goodfellow2020generative}, Energy Based Models (EBMs)~\cite{zhu1998filters,lecun2006tutorial,lu2016learning,xie2016theory} or their close relatives, Diffusion Models~\cite{sohl2015deep,ho2020denoising}.
The latter %which
are becoming increasingly popular nowadays due to their amazing ability to generate photorealistic images~\cite{dhariwal2021diffusion}.

EBMs offer several fundamental advantages over their competitors due to their simplicity:  A single neural network is involved in training, which means that fewer parameters need to be learned and training is less costly. 
They are also particularly appealing for interpretive applications: Once trained, the energy function can be analyzed with statistical mechanics tools~\cite{decelle2021restricted}, or shallow EBMs can serve as an effective model to ``learn" something from the data. EBMs have been exploited for instance to infer the three dimensional structure~\cite{morcos2011direct} or building blocks~\cite{tubiana2019learning} of proteins, to generate artificial pieces of genome~\cite{10.1371/journal.pgen.1009303}, for neuroimaging~\cite{HJELM2014245}, simulation of complex wavefunctions in quantum many-body physics~\cite{doi:10.1126/science.aag2302,melko2019restricted}, or to impute missing input~\cite{stairs}.
Furthermore, EBMs naturally implement compositionability,
i.e. the possibility of combining information from different trained models to generate new composite data~\cite{hinton1999product,du2020compositional}.

EBMs have been around for a very long time, starting with the well-known single-layer~\cite{ackley1985learning}, Restricted ~\cite{smolensky1986information} or the Deep~\cite{salakhutdinov2009} Boltzmann machines, to modern applications where EBMs are encoded in Deep Convolutional Neural Networks (ConvNets)~\cite{nijkamp2019learning,Nijkamp_Hill_Han_Zhu_Wu_2020,du2019implicit,du2020compositional}, which show very similar performance in image generation as state-of-the-art GAN methods~\cite{du2019implicit}. 
Despite their many advantages, the direct training of EBMs is hard and unstable in practice, a problem directly related to the approximations made to estimate the  negative term of the gradient~\cite{decelle2021equilibrium}.
%which is intractable and must be approximated by sampling methods, typically Markov Chain Monte Carlo (MCMC) or Langevin dynamics. Since such sampling is required each time parameters are updated, practical applications require using a number of iterations far shorter than the mixing time of the algorithm, the latter ensuring an ergodic exploration of the model phase space and a reliable estimation of the log-likelihood gradient.
As it is intractable, it must be approximated by sampling methods, typically Markov Chain Monte Carlo (MCMC) or Langevin dynamics, and such sampling is required each time parameters are updated.
However, practical applications use a number of iterations far shorter than the mixing time of the algorithm, required to ensure an ergodic exploration of the model phase space and a reliable estimation of the log-likelihood gradient.
To circumvent this ``equilibration'' problem, several strategies have been proposed, mainly relying either on good initializations of the MCMC chains to reduce the equilibration time, such as the (persistent) contrastive divergence method~\cite{hinton2002training,tieleman2008training} and its refinements~\cite{du2019implicit}, or on perturbative methods to approximate the gradient~\cite{gabrie2015training,kim2016deep,kumar2019maximum}.
\\
A very successful alternative to avoid altogether worrying about thermalization of the sampling processes during training %involved in the training of EBMs,
has been to give up encoding the empirical data statistics in the Boltzmann weight of a given energy model, as it is the normal EBMs' modus operandi.
%but adjusting
The idea is instead to adjust the parameters of that model to reproduce a dynamical path able to generate new data points~\cite{sohl2015deep}, using an idea inherited from out-of-equilibrium thermodynamics.
That is, sacrificing the usual equilibrium-like interpretability of the EBMs for the sake of a fast and high quality generation performance. This idea is at the core of the so-called diffusion models~\cite{ho2020denoising}.

Furthermore, recent experimental studies
in several EBM families have shown that such a dynamic path is also printed in EBMs when non-convergent samplings are used to train the models, and that this 
can be easily exploited to generate high-quality samples
in short times both in sampling and training~\cite{nijkamp2019learning,Nijkamp_Hill_Han_Zhu_Wu_2020,decelle2021equilibrium,muntoni2021adabmdca,liao2022gaussian,xie2022tale}.  It was shown  that this optimal regime for sample generation of EBMs is achieved by forcing out-of-equilibrium sampling of the model during training, specifically by always initializing the Markov chains with random conditions and iterating the MCMC process for only a few steps $k$.
The key observation was that particularly high-quality samples are generated when sampling the trained model for exactly the same time and initial distribution that was used to train the model~\cite{nijkamp2019learning,decelle2021equilibrium}. Nicely, these observations can be rationalized by moment-matching arguments at the fixed point of the training dynamics, as described in Ref.~\cite{nijkamp2019learning}. Furthermore, the apparition of this strong non-monotonic performance in the sample generation can be directly related to the MCMC mixing time, which typically increases with learning progress~\cite{decelle2021equilibrium}.
In fact,  the same work showed that most of the instabilities and difficulties typically encountered in training RBMs are directly related to this out-of-equilibrium regime, since all these effects disappear when the number of Gibbs steps used to compute the gradient exceeds the thermalization time, %for example,
e.g. when an incredibly long number of steps per iteration is considered~\cite{decelle2021equilibrium}. 

In this paper, we provide a first-principle analytical explanation of these previous experimental observations. 
In our theoretical framework, we show that the Gibbs-Boltzmann distribution of models trained in the out-of-equilibrium regime poorly reproduce the target statistics of the dataset. We also show that non-convergent samples can be used to accurately reproduce certain statistics of the dataset, but only through a very specific dynamic path fixed by the training choice.
We first prove these results for generic EBMs. Then, %In the following sections,
we analyze two simple EBMs, the Gaussian model and the Bernoulli Boltzmann machine in the early stages of training, where the effects introduced into the model by the non-convergent samplings can be analytically described along the entire training dynamics. 
Finally, we illustrate this phenomenon by performing real learnings in a ConvNet EBM and the  Boltzmann machine.

\section{Preliminaries: Maximum-likelihood training of Energy-Based Models}
\label{sec-preliminaries}

Suppose $\bx$ is the value taken by a data point.
%Let us now
We consider an arbitrary energy-based model (EBM) in order to model its probability in terms of a Gibbs-Boltzmann distribution:
\begin{equation}
\label{eq:ptheta}
p_{\bt} (\bx) = \frac{e^{-E_{\bt} (\bx)}}{Z(\bt)} \quad \text{with} \quad Z(\bt) = \int d\bx\ e^{-E_{\bt} (\bx)},
\end{equation}
where ${E_{\bt} (\bx)}$ is a dimensionless energy function that depends on a set of parameters ${\bt=\lbrace \theta_i \rbrace}$.
We let $E_{\bt} (\bx)$ be an arbitrarily complex function (e.g. parameterized by a ConvNet or a single-layer Boltzmann machine), the only requirement
being that $p_{\bt} (\bx)$ must exist together with its moments as functions of 
their conjugate variables $\bm{\theta}$.

Suppose now that we want to find a good set of parameters $\bt$ that minimizes the distance between $p_{\bt}(\bx)$ and the empirical distribution $\pemp(\bx)$ of a given set of training examples.
This is usually done by maximizing the log-likelihood:
\begin{equation*}
\like (\bt) = \av{\log p_{\bt}(\bx)}_{\pemp}=-\av{E_{\bt} (\bx)}_{\pemp}-\log Z(\bt),
\end{equation*}
where $\av{\cdot}_{p}$ denotes the average over a given distribution $p(\bx)$.
This function is intractable in general because so is $Z(\bt)$: as a consequence, $\like (\bt)$ is typically maximized through a (possibly stochastic) gradient ascent process. Its derivative with respect to parameter $\theta_i$ is
\begin{equation}
\label{eq:gradEBM}
\frac{\partial\like (\bt)}{\partial \theta_i} =\av{f_{i,\bt}(\bx)}_{ p_{\bt}}- \av{f_{i,\bt}(\bx)}_{\pemp},
\end{equation}
with the local `forces' in the phase space of parameters\\
\begin{equation}\label{eq:fdef}
f_{i,\bt}(\bx)\equiv\frac{\partial E_{\bt}(\bx)}{\partial \theta_i}.
\end{equation}
Note that $\av{f_{i,\bt}(\bx,\bt)}_p$ are referred as the generalized `moments' of the distribution $p(\bm x)$.
Similarly, the Hessian matrix  is given by the components
\begin{eqnarray*}
\label{eq:hessian}
\nonumber
& H_{ij}(\bt)
\equiv \frac{\partial^2\like (\bt)}{\partial \theta_i \partial \theta_j}
=\av{\frac{\partial f_{j,\bt}(\bx) }{\partial  \theta_i}}_{\! p_{\bt}} - \av{\frac{\partial f_{j,\bt}(\bx) }{\partial  \theta_i}}_{\!\pemp\!}
\\
&\!-\!\av{f_{i,\bt}(\bx) f_{j,\bt}(\bx)}_{\! p_{\bt}}\!
+\!\av{f_{i,\bt}(\bx)}_{\! p_{\bt}\!}\av{f_{j,\bt}(\bx)}_{\! p_{\bt}\!}\!.\!
\end{eqnarray*}
%\begin{eqnarray*}
%\label{eq:hessian}
%\nonumber
%& H_{ij}(\bt) =\frac{\partial^2\like}{\partial \theta_i \partial \theta_j} =\av{\frac{\partial f_{\theta_i}(\bx) }{\partial  \theta_j}}_{\!\pemp\!}-\av{\frac{\partial f_{\theta_i}(\bx) }{\partial  \theta_j}}_{\! p_{\bt}}\\
%&\!+\!\av{f_{\theta_i}(\bx,\bt) f_{\theta_j}(\bx,\bt)}_{\! p_{\bt}}\!-\!\av{f_{\theta_i}(\bx,\bt)}_{\! p_{\bt}\!}\av{f_{\theta_j}(\bx,\bt)}_{\! p_{\bt}\!}\!.\!
%\end{eqnarray*}
This means that in models where all $f_{j,\bt}(\bx)$ are a function only of $\bx$ and not of the $\bt$
(e.g. $E_\bt$ is a linear function of the parameters),
the optimization problem is always convex (i.e. the Hessian is negative semi-definite). This occurs because the covariance matrix is positive semi-definite. 
We will later analyze various models that satisfy this condition.

In practice the gradient \eqref{eq:gradEBM} is approximated by a stochastic sampling process (MCMC or Langevin dynamics). Thus, it is correctly estimated as long as this stochastic process converges to equilibrium.
However, such a sampling process must be repeated each time a gradient ascent step is performed.
As a consequence, ensuring convergence at each training step is not only tedious, but also infeasible in terms of the time required to train an adequate EBM.
A shortcut was proposed by Hinton: a few Gibbs sampling steps are sufficient if the sampling process is initialized wisely.
This is the idea underlying the recipe of contrastive divergence (CD), where the chains are always initialized at the training samples in the minibatch,
or of its refinement, persistent CD (PCD), where the configurations reached after a sampling process are kept as initial configurations for the subsequent sampling process after an update of the parameters.

The effects introduced into EBMs by the lack of convergence of these Markov chains at each update of the training, have gone largely unnoticed for decades, and only recently has it become clear that they introduce memory effects on the training process into models.
Even more interesting is the observation that
this issue can be exploited to our advantage:
non-persistent short MCMC runs are %indeed
the most efficient strategy to train good generative EBMs,
admittedly at the cost of forgetting the link between the equilibrium properties of the family of models parametrized by $\bt=\lbrace \theta_i \rbrace$, 
and the statistics of the dataset encoded in $\pemp$.
In the next section, we provide an analytical explanation of these results.

\section{Finite-time sampling: The master equation} \label{sec:master_eq}

The implications of using short, nonconvergent MCMC runs to estimate the gradient of Eq.~\eqref{eq:gradEBM} 
in the EBM training can be formalized and proved generically as follows. For simplicity, we assume here that the number of possible states of our model is finite and that time is continuous. The description for discrete time can be found in the Appendix~\ref{ap:discrete}, and the generalization to continuous variables in the Appendix~\ref{ap:contvar}. The latter case will also be considered later in an example.

We consider a MCMC sampling for a given model ${E_\bt}$, and denote $p_a(k)$ as the probability of being at the $a$-th possible configuration at time $k$. From now on, we denote the {\em sampling} time by $k$ to distinguish it from $t$, the {\em training} time.
Firstly, the {\em master equation} describes the time evolution of the probability of the system to occupy each of the states $a$
\begin{eqnarray}
\textstyle\frac{\der p_a(k)}{\der k} = \textstyle\sum_{b (\neq a)}w_{b \to  a} \, p_b{(k)}
-\sum_{b (\neq a)}w_{a \to  b} \, p_a{(k)},
\label{eq:mastereqcon}
\end{eqnarray}
where ${w_{a\to b}}$ is the transition probability of going from state $a$ to $b$ for that model. 
These transition probabilities are independent of $k$ as the process is by definition Markovian.
We introduce the vector notation $\bm p{(k)}$, whose $a$-th component is $p_a{(k)}$. The evolution of the probabilities can then be described by a {\em stochastic evolution} matrix $\bU$ ($\U_{ab}=w_{b \to a} \: \text{for } a\neq b$ and $\mathcal{U}_{aa}
=-\sum_{b (\neq a)} w_{a \to  b}$),
as
\begin{equation}
\label{eq:evolp}
\textstyle\dot{\bm p}{(k)}=\bU {\bm p}{(k)}
\quad \Rightarrow \quad
{\bm p}{(k)}
=e^{k{ \bU }}\,{\bm p}_0 ,
\end{equation}
where $\bm p_0$ represents the initial distribution (i.e. ${\bm p}(0)$).

Secondly, if the Markov chain is reversible, or equivalently satisfies detailed balance (which is the typical situation in EBMs), then the matrix $\bU$ is guaranteed to have a spectral decomposition.
Therefore, $\bm p(k)$ can be explicitly obtained from \eqref{eq:evolp} through a spectral expansion.
Let $\{\lambda_\alpha\}$ be the eigenvalues of $\bU $, and $\bm u_{\alpha}$ the associated eigenvectors.
Rewriting the initial distribution $\bm p_0$ in this basis,
we have
\begin{equation}
\bm p_0 =\textstyle\sum_\alpha c_\alpha {\bm u}_\alpha
\quad \Rightarrow \quad
{\bm p}(k)
=\textstyle\sum_\alpha c_\alpha e^{k \lambda_{\alpha}} \bm u_{\alpha},
\end{equation}
with the coefficient $\lbrace c_\alpha \rbrace$ encoding the memory of the initial condition of the sampling run.
The probability $\bm{p}{(k)}$ must be normalized at all $k$ (i.e. $\sum_a p_a{(k)}=1$), hence at least one eigenvalue must be $0$, and the rest must be necessarily negative for the long-time probabilities to remain bounded.

Thirdly, when $\bU$ is also irreducible (i.e. any configuration can be reached from any other in a finite number of steps), it follows from the Perron-Frobenius theorem that the $0$ eigenvalue must be unique (namely, $\lambda_0$)~\cite{keizer1972solutions}. We emphasize here that the theorem holds whether or not MCMC chains mix in a given simulation. It is a property that arises solely from the structure of the stochastic matrix.
Furthermore, one can always choose $c_0=1$,
which imposes from the long-time normalization that ${\sum_a u_{0,a}=1}$, and consequently for the initial normalization that all the eigenvectors with negative eigenvalues must fulfill $\sum_a u_{\alpha,a}=0$ with $\alpha\neq0$. Under these constraints, we have then
\begin{equation}
\label{eq:probev} 
\textstyle {\bm p}{(k)}=\textstyle\bm u_{0}+\sum_{\alpha>0} c_\alpha\, e^{-k| \lambda_\alpha|}\bm u_{\alpha}.
\end{equation}
An analogous expression is obtained for discrete sampling steps in Appendix~\ref{ap:discrete} and for continuous variables for the states in Appendix~\ref{ap:contvar}.
By construction,  it is guaranteed that the stochastic dynamics converges to the distribution $\bm u_0$ after a long time. Moreover, the {\em mixing} time is $\kappa_\mathrm{mix}\!=\!1/| \lambda_1|$, with $\lambda_1$ the second smallest absolute eigenvalue. Up to this point, the discussion was completely generic.

Fourthly, we can now explicitly focus on EBMs by forcing the long-time distribution ${\bm u}_0$ to coincide with the Gibbs distribution~\eqref{eq:ptheta}, i.e. $u_{0,a}= p_{\bt}(\bx^{(a)}) \equiv \exp\caja{-E_\bt (\bx^{(a)})}/Z(\bt)$, where  $\bx^{(a)}$ are the  variables in configuration $a$.
To be consistent, ${\bm u}_0$ must be the solution of the steady-state condition $\bU (\bt)\,\bm p_{\bt}=\bm 0$ (cf.~\eqref{eq:evolp}). It is in particular directly satisfied if we impose the {\em detailed balance} condition $w_{a \to  b}\, e^{-E_\bt(\bx^{(a)})}=w_{b \to  a}\,e^{-E_\bt(\bx^{(b)})}$
%\begin{equation*}
%    \frac{w_{a \to  b}}{w_{b \to  a}}=e^{E_\bt(\bx^{(a)})-E_\bt(\bx^{(b)})},
%\end{equation*}
to hold, which has previously been imposed for the spectral expansion to exist. Standard choices for the update rules, such as the Metropolis-Hastings or the Heat Bath dynamics satisfy detail balance.

Finally, under this choice of transition probabilities, and the absence of some pathological cycles in the dynamics, we have the guarantee that the sampled 
distribution will converge to the equilibrium one with increasing $k$ as
\begin{equation} 
  {\bm p}{(k)}=\bm p_{\bt}+\bm \Lambda_{\bt}\paren{k,\bm p_0},
  \label{eq:dynU}
\end{equation}
where $\bm\Lambda_\bt$ is the term $\sum_{\alpha > 0}$ in Eq.~\eqref{eq:probev}, a decay function of $k/\kappa_\mathrm{mix}$ at long times.
One cannot compute this function $\bm\Lambda_\bt$ in general, but we know it depends only on
the initial conditions $\bm p_0 \leftrightarrow \lbrace c_\alpha \rbrace$, the run duration $k$, and the evolution matrix ${\bU (\bt)}$. Then, the expected value for $\bm f_{\bt}(\bx)\equiv\bm\nabla E_{\bt}(\bx)$ in the log-likelihood gradient (cf.~Eqs.~\eqref{eq:gradEBM}-\eqref{eq:fdef}), estimated after a sampling time $k$, is given by% (noting $f_{i,\bt}(\bx^{(a)})$ as $f_{i,\bt}^{(a)}$)
\begin{eqnarray}
\label{eq:finitek}
\av{\bm f_{\bt}}_{k, \bm p_0}&\!\equiv\!&\textstyle\sum_a \bm f_{\bt}^{(a)}\, p_a {(k)}\\\nonumber  &\!=\!& \av{\bm f_{\bt}}_{ p_{\bt}}\!+\!\textstyle\sum_a \bm f_{\bt}^{(a)}\Lambda_{\bt,a}\paren{k,\bm p_0}.
\end{eqnarray}
with $f_{i,\bt}^{(a)} \equiv f_{i,\bt}(\bx^{(a)})$. In other words, by approximating the gradient with a non-convergent MCMC sampling, one computes the real model average plus a correction depending on $k$, $\bm p_{0}$, and the model parameters $\bm \theta$.
Thus, at each training update, the update rule of Eq.~\eqref{eq:gradEBM} is replaced by 
\begin{eqnarray*}
\label{eq:gradEBM-fink}
\left.\frac{\partial\like (\bt)}{\partial \theta_i}\right|_{k,\bm p_0}\!=\!\av{ f_{i,\bt}}_{ p_{\bt}\!}\!-\!\av{ f_{i,\bt}}_{\pemp\!}\!+\!\sum_a\! f_{i,\bt}^{(a)}\textstyle\Lambda_{\bt,a}\!\paren{k,\bm p_0}.
\end{eqnarray*}
Moreover, this implies
that the fixed point $\bt=\bt^*$ of this dynamics (i.e. $\nabla \mathcal{L}=0$), if it can be reached,
satisfies
\begin{equation}
\label{eq:fixed-point-EBM}
    \av{\bm f_{\bt^*}}_{\pemp}\!=\!\av{\bm f_{\bt^*}}_{ p_{\bt^*}}\!+\!\textstyle\sum_a \bm f_{\bt^*}^{(a)}\Lambda_{\bt^*\!,a}\!\paren{k,\bm p_0}.
\end{equation}
\vspace{-0.5cm}
\begin{theorem}
    \label{thm:markov}
    We consider %Assuming
    a training of an EBM that follows the gradient in Eq.~\eqref{eq:gradEBM} estimated with nonconvergent, irreducible, and reversible Markov chains.
    If it reaches a dynamic fixed point where $\bm \nabla \mathcal{L}=\bm 0$, then, the trained model generates samples consistent with the statistics of the training set, i.e. same generalized moments -- given by $\av{\bm\nabla E_{ \bt^*}}_{k,\bm p_0}\!=\!\av{\bm\nabla E_{ \bt^*}}_{p_\mathcal{D}}$ --, but only if the samples are drawn via the same identical stochastic process used for the training: i.e., the same update rules, the same initialization, and the same Markov chain length.
\end{theorem}
The proof results from combining the Eqs.~\eqref{eq:finitek}-\eqref{eq:fixed-point-EBM} using a model with parameters ${\bt^*}$ trained with a given sampling dynamics and $(\bm{p}_0,k)$. Now, the samples generated with this model, with ${\bU({\bt^*})}$, an arbitrary initial distribution ${\bm q_0}$ and after a time $k^\prime$, must satisfy the following statistics:
\begin{align}
\label{eq:errorfinitek}
&\av{\bm f_{\bt^*}}_{k^\prime \! , \bm q_0}
=\av{\bm f_{\bt^*}}_{\pemp}+\bm{D}_{\bt^*}(k^\prime,\bm q_0,k,\bm p_0),& 
&
\end{align}
%\vspace{-0.75cm}
with the mismatch given by Eq.~\eqref{eq:probev}:
\begin{align*}
    & %\text{with}\:
    \bm{D}_{\bt^*\!}(k^\prime,\bm q_0,k,\bm p_0)\!
    = \! \textstyle\sum_{a} {\bm f}_{\bt^*\!}^{(a)} \left[ \Lambda_{\bt^*\!,a}(k^\prime \!,{\bm q}_0)\!
    -\! \Lambda_{\bt^*\!,a} (k,{\bm p}_0) \right]
    \\
    &\!=\!\textstyle\sum_{a,\alpha}\! \bm f_{\bt^*}^{(a)} \paren{c_\alpha({\bm q_0})e^{-k^\prime|\lambda_\alpha|}\!- \! c_\alpha(\bm p_0)e^{-k| \lambda_\alpha|}} u_{\alpha,a}.&
\end{align*}
and the model-dependent $\lbrace \lambda_\alpha (\bt^*),\bm u_\alpha (\bt^*) \rbrace$.
This means that if we choose the same sampling time ${k^\prime \!=k}$,
the same initialization probabilities ${\bm q_0= \bm p_0}$,
and the same dynamic rules as used in learning, then we will match perfectly the empirical statistics. This statement holds for all derivatives of the energy function $E_{\bt^*}$ with respect to the model parameters.
Also, a long sampling ($k^\prime\!\to\!\infty$) can deviate strongly from the empirical expectation if the training $k$ is much shorter than the mixing time $\kappa_{\text{mix}}$. This discrepancy results from the remaining factors ${\lbrace -c_\alpha({\bm p}_0) \, \exp^{-k \vert \lambda_\alpha \vert} \rbrace}$ in Eq.~\eqref{eq:errorfinitek}, which would vanish in the opposite limit $k \!\gg\! \kappa_{\text{mix}}$.

The results of Theorem~\ref{thm:markov} also hold before reaching the fixed point [i.e. for trained nonconvex EBMs or unfinished trainings of convex EBMs] but with several limitations. 
\begin{theorem}
    \label{thm:markovbis}
    We consider a non-convergent EBM trained
    to a point where $\bm\nabla\mathcal{L} (\bt^\dagger)=\bm\varepsilon \neq \bm 0$.
    Then, the samples generated by the model $\bt^\dagger$ (with the same stochastic process used for training --the same $k$, the same initialization $\bm p_0$ and update rules) reproduce the statistics of $\av{\bm\nabla E_{ \bt^\dagger}}_{p_\mathcal{D}}$ of the training set with an error $\bm\varepsilon$. Thus, the error committed after time $k$ decreases (in average) as the training approaches its fixed point. Moreover, for a conveniently small $\bm \varepsilon$ (which depends on $\bm p_0$, $k$ and $\bm f_{\bt^\dagger}$), it is possible to find a sampling time $k^\dagger( \varepsilon_i)\neq 0 $ at which the error for $\av{f_{i,\bt^\dagger}}_{p_\mathcal{D}}$ is zero. This $k^\dagger$ moves toward the length $k$ of the Markov chains used for training as the training approaches its fixed point.
\end{theorem}
This second theorem is straightforward from the definition of the gradient and Eq.~\eqref{eq:errorfinitek}.
Then, upon  sampling model $\bt^\dagger$ with the same stochastic matrix $\bU (\bt^\dagger)$, the absolute error of $f_{\bt{^\dagger}}$ averaged on samples generated at a time $k^\prime$ from the initial distribution $\bm{q}_0$ is:
\begin{eqnarray*}
\textstyle\left|\av{\bm f_{\bt^{\dagger}}(\bx)}_{k^\prime,\bm p_0} \! -\av{\bm f_{\bt^\dagger}(\bx)}_{\pemp}\right| \! =\textstyle\left|\bm\varepsilon + \bm{D}_{\bt^\dagger}(k^\prime,\bm q_0,k,\bm p_0)\right|,
\end{eqnarray*}
and $\bm{D}_{\bt^\dagger}(k^\prime,\bm q_0,k,\bm p_0)$ vanishes for $k^\prime=k$ and $\bm q_0=\bm p_0$. The second statement stands from the fact that $\bm{D}_{\bt^\dagger}$ is a smooth function of $k^\prime$ that (for $k$ short compared to $\kappa_\mathrm{mix}$) changes sign around $k^\prime=k$.
This means that the error of $\av{f_{i,\bt}}_\pemp$ will be strictly zero at $k^\dagger(\varepsilon_i)\neq k$ for $\varepsilon_i$ sufficiently small,  with $\varepsilon_i$ being the discrepancy between the statistics of the data set and the model associated to the $i$-th moment statistics). This means that $k^\dagger$ moves progressively toward $k$ as $\varepsilon_i \to 0$. In a typical MCMC scheme, time is always an integer variable, so this optimum might not be easy to access. However, in Langevin dynamics this is possible, as we show in a later example. This also means that as learning progresses ($\bm \varepsilon\ \to \bm 0$), the best sampling time $k^\dagger$ progressively converges to $k$ for all `moments' simultaneously.

We assess sample quality based on the agreement of the energy moments, $\av{f_{i,\bt}}$, between the training and the generated sets. The key question here is whether the samples resulting from convergent MCMC training are more representative of the distribution of the dataset than those obtained using a non-persistent, non-convergent strategy. The answer is not straightforward. An immediate conclusion from previous results is that convergent MCMC training does not improve the agreement of moments because $\bm p_\bt$ reproduces the same statistics than $\bm p(k,\bm p_0)$ in the nonconvergent protocol. The same is true for all observables that can always be expressed as functions of the distribution's `moments' \eqref{eq:fdef}. However, no further generalizations can be made for the rest of observables because the expected committed error is not only related with the non-convergent sampling, but also with the limitations of the model chosen (which may not be powerful enough to capture all features of the dataset), and both contributions can have any sign. Consequently, these changes may either cancel or accumulate, depending on the observable or EBM under consideration.

\begin{figure*}[h!]
    \centering\includegraphics[height=5cm,trim=0 20 0 0]{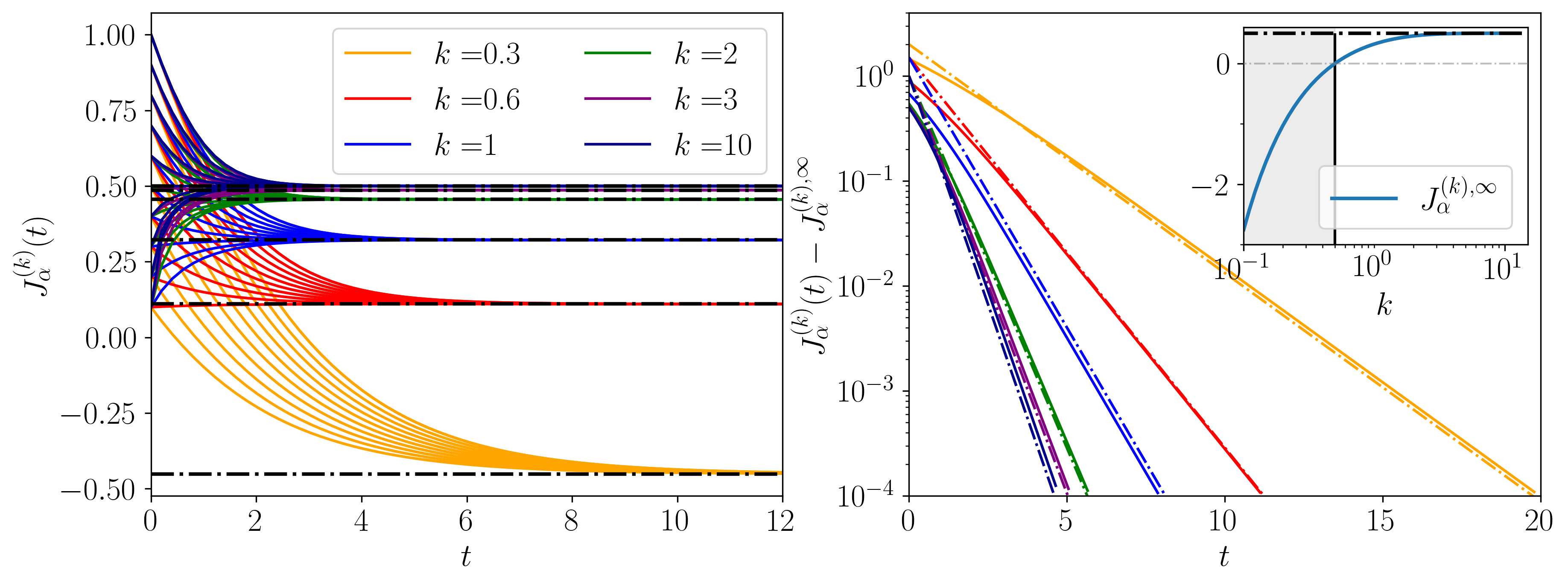}
    \caption{\textbf{Left:} Evolution of $J_\alpha^{(k)}(t)$ for different values of $k$ (different colors) and initial conditions for $J_\alpha^{(k)}(0)$. \textbf{Right:} Convergence for a large training time $t$, where the higher $k$, the faster the convergence; dash-dotted lines are the numerical integration of Eq.~\eqref{eq:grad_Jak}, full lines show the exponential fit $J_\alpha^{(k)} (t) - J_\alpha^{(k),\infty}\! \sim\! A\exp(-t/\tau)$ with $\tau$ given by Eq.~\eqref{eq:taugauss}. \textbf{Inset:} the asymptotic values $J_\alpha^{(k)}(t\!\to\!\infty)$ as a function of $k$. The gray shaded region corresponds to the non-physical solutions and the vertical line to the limit $k = (\hat{c}_{\alpha \alpha} - \av{x_\alpha^2}_0)/2$.}
    \label{fig:EvolFiniteKInset}
\end{figure*}
\section{Solvable models}
In the next sections we will examine two simple classes of EBMs for which:
\textit{(i)}~the correction
$\bm \Lambda_{\bt}(k,\boldsymbol{p}_0)$
introduced by the non-convergent Markov chains can be explicitly computed, and \textit{(ii)}~its effect on the inferred model parameters can be explicitly quantified along the entire learning process.
Specifically, we focus on the inverse problem for families of exponentials with quadratic arguments:
Given a dataset drawn from an empirical distribution $\pemp(\bx)$, we wish to infer the best parameters  ${\bt}\equiv\{\bJ,\bh\}$ of the target distribution
with an energy function of the type
\begin{align*}
    %p_{\bt} (\bx) &= \frac{1}{Z(\bt)}e^{-E_{\bt} (\bx)}, \\ 
    \textstyle E_{\bt} (\bx)&=\frac{1}{2} \textstyle\sum_{ij} x_i J_{ij} x_j + \sum_i h_i x_i.
    %Z(\bt) &= \int d\bx\ p_{\rm{prior}}(\bx) e^{-E_{\bt} (\bx)},
\end{align*}
%So far, we postpone the necessary conditions for these parameters to ensure that the target distribution is well-defined. 
For simplicity we consider $h_i=0$ in the following, but a generalization with external fields/biases is straightforward.
As in the general case in Sec.~\ref{sec-preliminaries}, we estimate the parameters by maximizing the log-likelihood of the model evaluated on the dataset, which reads
%where in this model reads 
\begin{equation}
    \like (\bt) =\textstyle -\frac{1}{2} \sum_{ij} J_{ij}\av{x_i x_j}_{\pemp} - \log{Z(\bt)}.
\end{equation}
Again, the classical approach to maximizing log-likelihood is based on calculating its gradient
\begin{align}
    \frac{\partial \like (\bt)}{\partial J_{ij}} &= -\paren{\av{x_i x_j}_{\pemp} - \av{x_i x_j}_{p_{\bt}} }\!, \label{eq:gradJ}
\end{align}
The average over ${\bm p}_\bt$
is usually calculated using the heat-bath MCMC algorithm. In the next subsections, we analytically investigate the effect of replacing the equilibrium averages with estimates obtained by a non-convergent MCMC process.
We  first consider the Gaussian model and then the high-temperature regime of the inverse Ising problem.

\subsection{Exact results for the Gaussian model}

In the Gaussian case, we consider data points $\bx \in \mathbb{R}^D$. To be well-defined, the coupling matrix $\bJ$ must be positively definite, namely all its eigenvalues must be strictly positive. Under this condition, and given the convexity of this likelihood function, we can explicitly compute the correlation function, and the stationary solution of Eq.~\eqref{eq:gradJ} reads:
\begin{equation*}
    \av{\bm x \bm x^{\text{T}}}_{p_{\bt}}= \bm J^{-1}
    \; \stackbin[k\to\infty]{[\nabla \mathcal{L}=0]}{\Longrightarrow} \;
    %\; \text{ giving } \;
    \bm J^{*} = \av{\bm x \bm x^{\text{T}}}_{\pemp}^{-1} .
\end{equation*}
%This solution must be found by the gradient ascent method, with a suitable  learning rate. 
Despite the maximum likelihood point can be found analytically, we now  compute it through a gradient ascent method, with a suitable learning rate. Moreover, the gradient can be projected on the eigenbasis of $\bJ$. Denoting {$\lbrace \bm{u}^\alpha \rbrace$} the eigenvectors, we have that $J_{ij} = \sum_\alpha u_i^\alpha J_\alpha u_j^\alpha$. Following~\cite{decelle2017spectral,decelle2018thermodynamics,decelle2021restricted}, we can write a set of evolution equations for the eigenvalues {$\lbrace J_\alpha \rbrace$} and for the rotation {$\Omega_{\alpha \beta}$} of the eigenvectors
\begin{align}
    \frac{\der J_\alpha}{\der t} &= -\hat{c}_{\alpha \alpha} + 1/J_\alpha, \label{eq:grad_Ja}\\
    \Omega_{\alpha \beta} & {\equiv} \bm{u}^\alpha \frac{\der\bm{u}^\beta}{\der t} = \frac{\langle x_\alpha x_\beta \rangle_{p_\bt} - \hat{c}_{\alpha \beta}}{J_\alpha - J_\beta}, \label{eq:grad_ua}
\end{align}
where we defined $\hat{c}_{\alpha \beta} = \sum_{i,j} u_i^\alpha u_j^\beta \av{x_i x_j}_{\pemp}$ and $x_\alpha = \sum_i x_i u_i^\alpha$, the projection on the eigenmodes of $\bJ$ of the correlation matrix and $\bx$, respectively.
In this formulation, we see that the eigenvectors have to diagonalize the correlation matrix to cancel the rotation,
i.e. each vector 
$\bm{u}^\alpha$ will align itself with the principal directions of the dataset, and
the eigenvalues will eventually adjust to $J_\alpha = 1/\hat{c}_{\alpha \alpha}=J_\alpha^*$.

However, we are interested here in the case where non-convergent MCMC chains are used to approximate the correlation function $\av{\bm x \bm x^{\text{T}}}_{p_{\bt}}$, and what it changes for Eqs.~\eqref{eq:gradJ}-\eqref{eq:grad_Ja}-\eqref{eq:grad_ua}. We assume for simplicity that the dataset is centered with zero mean, $\av{x_i}_{\pemp} = 0$.
The evolution of the chain is described by the following Langevin equation
\begin{equation}\label{eq:gradientk}
    \frac{\der x_i(k)}{\der k} \!=\! -\!\textstyle \sum_j J_{ij} x_j (k)\! + \!\eta_i(k),% \,\,
\end{equation}
with $\bm \eta$ a Gaussian white noise of zero mean and variance $\overline{\eta_i (k) \eta_j(k')} = 2 \delta_{ij} \delta(k-k')$, where $\overline{(\cdot)}$ denotes the average with respect to different independent trajectories. These equations can be decoupled by projecting them on the eigenmodes of $\bJ$, leading upon sampling to
%to the following solution
\begin{equation*}
    x_{\alpha}(k)=x_{\alpha}(0)e^{-J_\alpha k}+\int_0^k \mathrm{d}\tilde{k} \ e^{J_\alpha (\tilde{k}-k)}\eta_\alpha(\tilde{k}).
\end{equation*}
We see that the convergence properties are dominated by the eigenvalues $J_\alpha$. Hence, for $k \lesssim J_\alpha^{-1}$, the MCMC chain will be far away from its equilibrium value. We can obtain a similar result for the correlation function; let us directly look at the average over realizations at time $k$%$t$
\begin{equation}
    \av{x_{\alpha}x_{\beta}}_{k,\bm p_0} \!=\!\av{x_\alpha x_\beta}_0 e^{-(J_\alpha+J_\beta)k}\!+\!\frac{\delta_{\alpha\beta}}{J_\alpha}\paren{ 1\!-\! e^{- 2 J_\alpha k}}, \label{eq:corrLangeving}
\end{equation}
where $\av{\cdot}_0$ is the average over the initialization distribution. Thus, the asymptotic value is given by $\av{x_{\alpha}x_{\beta}}_\infty = \delta_{\alpha \beta}/J_\alpha$.
\\
We now analyze how the learning dynamics are affected when the model average $\av{x_{\alpha}x_{\beta}}_{p_\bt}$ is replaced in the gradient by $\av{x_{\alpha}x_{\beta}}_k$, i.e. the averages computed at fixed time $k$ when the chains are initialized at a fixed random initial distribution $p_0(\bx)$. We first consider only the evolution of the eigenmodes given by Eq.~\eqref{eq:grad_Ja}, and keep the eigenvectors $\bm{u}^\alpha$ fixed. Under these conditions, the gradient calculated with non-convergent chains is quite different to the real one
\begin{align*}
    &\left.\left(\frac{\partial \mathcal{L}}{\partial \bm{J}}\right)_{\alpha\beta}\right|_{k,{\bm p}_0} ={-\left[\left\langle x_\alpha x_\beta \right\rangle_{\pemp}-\av{x_{\alpha} x_{\beta}}_{{k,\bm p_0}} \right]} 
    \\
    &=\left(\frac{\partial \mathcal{L}}{\partial \bm{J}}\right)_{\alpha\beta} + \paren{\av{x_\alpha x_\beta}_0 - \frac{\delta_{\alpha \beta}}{J_\alpha}}e^{-\paren{J_\alpha+J_\beta} k}.
\end{align*}
Note that the gradient now depends on both the sampling time
%number of sampling steps
$k$ and the initialization ${\bm p}_0$ of the chains.
Then the evolution for the eigenvalues of $\bm J$ in Eq.~\eqref{eq:grad_Ja} in the finite $k$ case transforms into
\begin{equation}\label{eq:grad_Jak}
    \frac{\der J_\alpha^{(k)}}{\der t}\! =\! -\hat c_{\alpha\alpha} + \frac{1}{J_\alpha^{(k)}}  \!+\!  \left(\av{x_\alpha^2}_0 \! -\! \frac{1}{J_\alpha^{(k)} }\right) e^{-2 J^{(k)}_\alpha k},
\end{equation}
whose stationary solution at ${t \to \infty}$, $J_\alpha^{(k),\infty}$, satisfies 
\begin{equation}
    \hat{c}_{\alpha \alpha} \!-\! \av{x_\alpha^2}_0 e^{- 2 J_\alpha^{(k),\infty} k}\! =\! \frac{1}{J_\alpha^{(k),\infty} }\! \left( 1\!-\!e^{-2 J_\alpha^{(k),\infty} k} \right). \label{eq:fixedpoint_gauss}
\end{equation}
\begin{theorem}
    \label{thm:conv_gauss}
    If $k > (\hat{c}_{\alpha \alpha} - \av{x_\alpha^2}_0)/2$ holds, then Eq.~\eqref{eq:fixedpoint_gauss} admits a strictly positive solution. Otherwise, $J_\alpha^{(k),{\infty}}$ is negative and the non-convergent learning procedure diverges.
\end{theorem}
This theorem is %easily
proved by analyzing the function $g(J) = J \hat{c}_{\alpha \alpha}-1 + e^{- 2 J k}(1-J\av{x_\alpha^2 }_0)$, see the appendix~\ref{ap:thm_k} for more details. Therefore, we can define a lower bound for the sampling time $k$ below which the learning dynamics leads to an unphysical solution (unbounded Gaussian integrals).
From Eq.~\eqref{eq:fixedpoint_gauss}
we recover straightforwardly
%it is also clear how
the correct solution $J_\alpha^* = 1/\hat{c}_{\alpha \alpha}$ %is recovered
in the limit $k \to \infty$.

\begin{figure*}[h!]
    \centering
    \includegraphics[height=5.cm,trim=0 20 0 0]{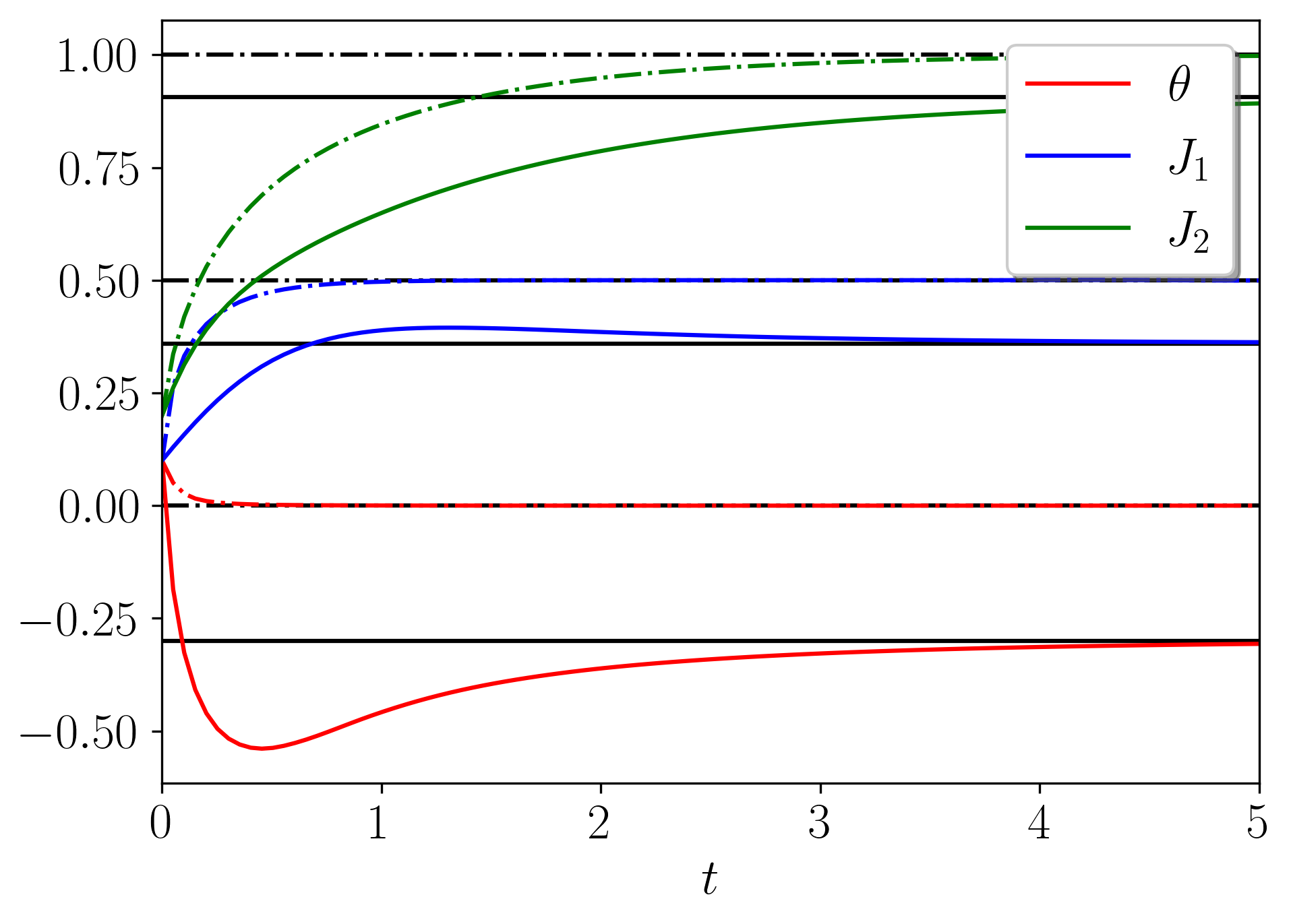}
    \includegraphics[height=5.cm,trim=0 20 0 0]{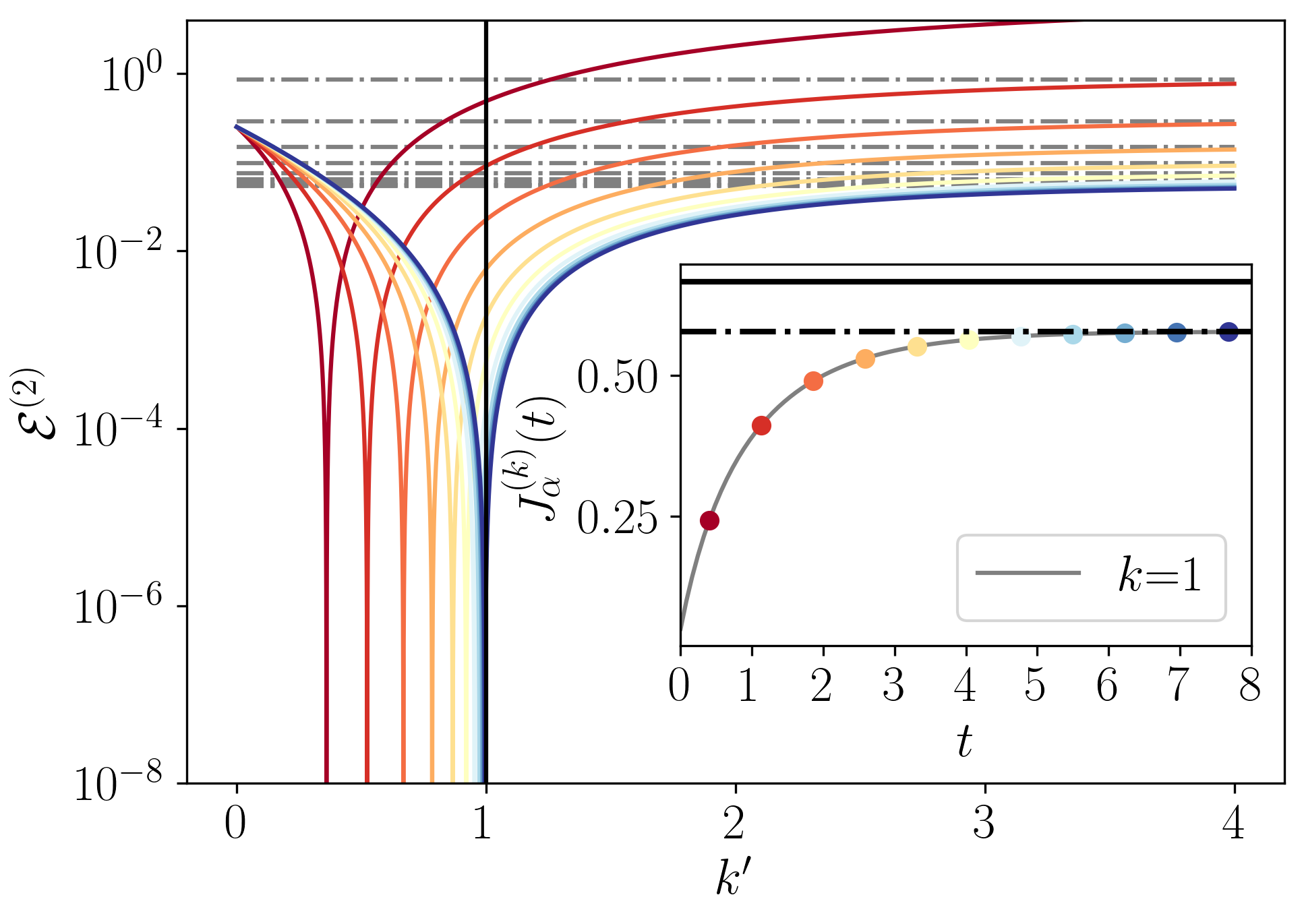}
    \caption{\textbf{Left:} Numerical resolution of the learning dynamics in presence of two modes. The dash-lines represent the resolution using a convergent dynamics $k \to \infty$, while the plain ones correspond to $k\!=\!1$. \textbf{Right:} \textbf{(inset)} evolution of $J_\alpha^{(k)}(t)$ for $k\!=\!1$ at various stages of the learning.  \textbf{(Main)}  The error on the correlation function, $\mathcal{E}^{(2)}\!=\!\textstyle(\av{x_\alpha^2}_{k^\prime,\bm p_0}\!-\!\hat{c}_{\alpha\alpha})^2$, of the machine at different stages of the learning. We can clearly see the peak of best performance (of several orders of magnitude) at a $k^{\dagger}$ that moves to $k$ as the learning converges. The color lines match the circles on the inset. The horizontal dashed-line represent the $k'\!\to\!\infty$ asymptotic value of the error.}
    \label{fig:EvolJResampling}
\end{figure*}

Equation~\eqref{eq:grad_Jak} can be integrated numerically, and the results for various $k$ and initial conditions are shown in 
Fig.~\ref{fig:EvolFiniteKInset} (left).
%It is clear that
The smaller $k$, the slower the convergence to the fixed point $J_\alpha^{(k),\infty}$.
Indeed, one can %easily
show that the typical training time $\tau$ required to reach the stationary point decreases with $k$ as 
\begin{equation}
\label{eq:taugauss}
\frac{1}{\tau} \!\sim\! \textstyle k\caja{\av{x_\alpha^2}_0 \!-\!\frac{ e^{-2k J_\alpha^{(k),\infty}}}{J_\alpha^{(k),\infty}}} \!+\! \frac{\left(1-e^{-2k J_\alpha^{(k),\infty}} \right)}{2\left(J_\alpha^{(k),\infty}\right)^2},
\end{equation}
by simply assuming an exponential relaxation for a large training time $t$, i.e., $J_\alpha^{(k)} (t)\! \sim\! J_\alpha^{(k),\infty} + A\exp(-t/\tau)$ in Eq.~\eqref{eq:grad_Jak}. Therefore, from a computational point of view, a smaller value of $k$ means by definition faster sampling, but the time required for the learning process to converge increases as $k$ decreases. Thus, the optimal value of $k$ results from a trade-off between these two contributions.

One can also study the effect of the non-convergent chains on the alignment of the eigenvectors of $\bm J$,
which we kept fixed until now.
For this, we rely on the simplified situation proposed in~\cite{decelle2021restricted}, where a reduced 2D subspace is considered in which the 2 first eigenvectors of $\bJ$ form an angle $\phi$ with the 2 principal directions of the dataset. In this case, one obtains a set of $3$ dynamical equations for $\phi(t)$, and the 2 eigenvalues $J_1(t)$ and $J_2(t)$
\begin{align*}
    \textstyle\frac{\der\phi}{\der t}  &\!=\! \textstyle\left[\sin(2\phi) (\hat{c}_1\!-\!\hat{c}_2)/2 \!+\! \av{x_1 x_2}_0 e^{- (J_1\!+\!J_2) k}\right]\!/\!(J_2\! -\! J_1),\\
    \textstyle\frac{\der J_1}{\der t} &\!=\! \textstyle J_1^{-\!1\!} \!-\! \hat{c}_1\! \cos^2\!(\phi) \!-\! \hat{c}_2\! \sin^2\!(\phi) \!-\! \left(J_1^{-\!1}\!-\!\av{x_1^2}_0\right) e^{\!-\!2 J_1 k}, \\
    \textstyle\frac{\der J_2}{\der t} &\!=\! \textstyle J_2^{-1\!}\! -\! \hat{c}_1\! \sin^2\!(\phi)\! -\! \hat{c}_2\! \cos^2\!(\phi) \!-\! \left(J_2^{-\!1\!}\!-\!\av{x_2^2}_0\right) e^{\!-\!2 J_2 k}.
\end{align*}
(see Appendix~\ref{ap:rot} for details) whose numerical solution is given in Fig.~\ref{fig:EvolJResampling}--left. 
In the limiting case $k\!\to\!\infty$, we see as expected the eigenvectors align (or anti-align) with the eigenvectors of the covariance matrix of the dataset and  $\phi\!\to\! 0 \text{ or } \pi$.
Yet, for finite $k$ this is not true, and an asymptotic finite $\phi$ is reached which embodies the sampling mismatch.

In Sec.~\ref{sec:master_eq}, we showed that generic EBMs trained with non-convergent MCMC samplings can be used to generate %draw
samples that perfectly match the statistics of the dataset even before the learning has converged.
For that purpose, we only need to mimic the same dynamic process used for training in the generation phase. We can then %easily
quantify this prediction in this solvable case,
given the %learnt
parameters of the model $J_\alpha^{(k)}(t)$ after a learning time $t$,
by computing the average correlations
through Eq.~\eqref{eq:corrLangeving}.
\begin{theorem}
     \label{thm:ReSamplingK}
     We consider a Gaussian model trained with non-convergent Markov chains of length $k$ along the modes $\alpha$ (i.e. keeping fixed the eigenvectors $\lbrace \bm{u}^\alpha \rbrace$), and parameterized by $\lbrace J_{\alpha}^{(k),\infty} \rbrace$. It
     exactly reproduces the correlations of the dataset when sampled with the same dynamic process used in the training: same transition rules, initialization distribution, and chain length.
\end{theorem}
The proof is straightforward by inserting the fixed point equation~\eqref{eq:fixedpoint_gauss} in~\eqref{eq:corrLangeving}. In particular, let us generate new data $\bm y$ with the trained model. %following a sampling process. It is clear that
When $k'=k$, the averaged correlation along  each mode $\alpha$
\begin{equation*}
    \av{y_\alpha^2}_{k,\bm p_0}=\hat{c}_{\alpha \alpha} + \left( \av{y_\alpha^2 }_0 -  \av{x_\alpha^2 }_0\right)e^{-2 J_\alpha^{(k),\infty} k},
\end{equation*} match those of the dataset but only if the chains are initialized from the same distribution used during the training, i.e, same $\av{x_\alpha^2 }_0$. In Fig.~\ref{fig:EvolJResampling}--right, we show the mean squared error $\mathcal{E}^{\left(2\right)}$ in the correlations committed at different stages of learning, i.e., using the numerically integrated $J_\alpha^{(k)}(t)$. As anticipated in Sect. 3, one can also reproduce the statistics of the dataset before convergence, but at an effective sampling time $k^{\dagger}$ that moves to $k$ as learning progresses.

As a final remark, from Eq.~\eqref{eq:corrLangeving} we see that the cross-correlation terms $\av{x_\alpha x_\beta}_{k}$ for $\alpha \neq \beta$ are zero only when the initial conditions are de-correlated along both modes, and thus the eigenvectors $\bm{u}^\alpha$ easily align with the principal directions of the dataset. In purely random initializations and to some extent in the PCD recipe, the cross terms are zero, but in the CD scheme this is not the case in the initial stages of learning, which should affect the rotational dynamics of the eigenvectors of the coupling matrix.
\begin{figure*}[t!]
    \centering 
    \includegraphics[trim=0 10 0 0,clip,width=\textwidth]{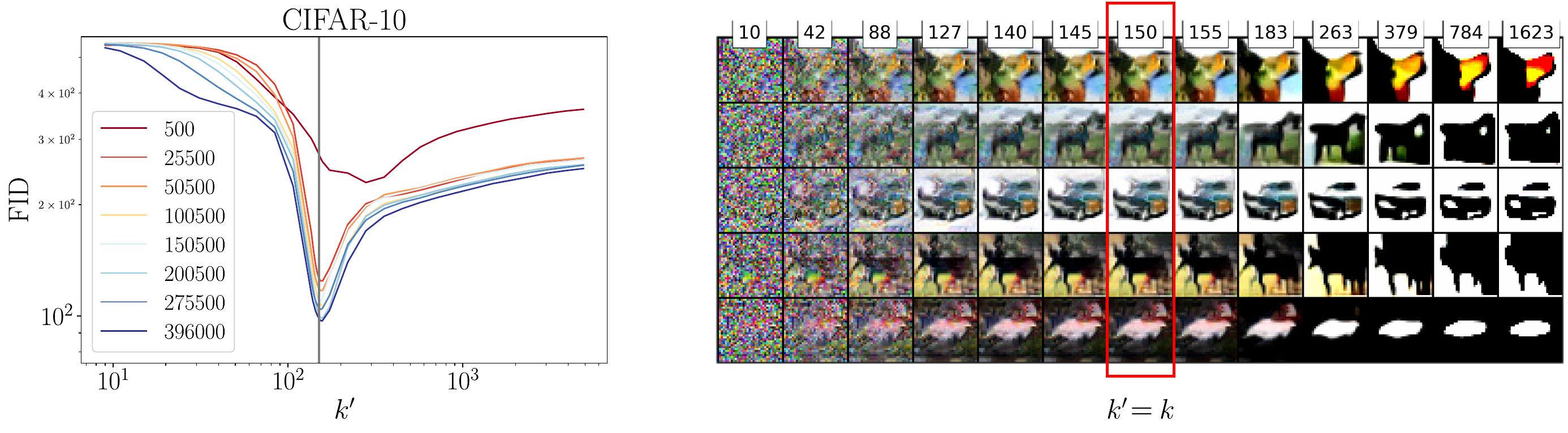}
    \vspace*{-7mm}
    \caption{ Generation results obtained with a ConvNet EBM trained on CIFAR-10 using $k=150$ Langevin MCMC sampling steps from random initial conditions. \textbf{Left:} We use the Frechet Inception Distance Score (FID) to evaluate the generation quality as a function of the sampling time $k^\prime$. The different colors correspond to different training epochs. We see again that the best score is achieved at $k'=k$ (corresponding to the gray line). \textbf{Right:} Example of $5$ generated images as a function of $k'$ for the most trained machine.}
    \label{fig:rescifar}
\end{figure*}

\subsection{Exact results on the Boltzmann machine}
The previous results can be extended to the case of binary variables, such as the classical Ising inverse problem~\cite{zecchinareviewinverseising}, or the Boltzmann machine (BM). In this context, we have $\bx = \{\pm 1\}^N$, for which the distribution is defined for all $\bJ$. In general, the learning of the BM cannot be solved exactly, since there is no general closed-form expression for the correlation function. However, we can deal with the so-called mean-field case (MF), where the coupling matrix scales with $1\slash N$. In this regime, BM can be solved analytically using %linear response theory in many MF regimes
MF theory~\cite{10.1162/089976698300017386,Nguyen_2012,Ricci-Tersenghi_2012}.
It is then possible to derive the evolution of the magnetization and the correlation function for a given MCMC dynamics. %quite easily.
Following~\cite{suzuki1968dynamics}, we first define the heat-bath dynamics transition matrix as the probability of a given spin $s_j$ when all others are fixed: $W_j(s_j^{t+1} | \{s_i^{t}\}_{i \neq j})$. Then, using the master equation as Eq.~\eqref{eq:mastereqcon}, it is possible to write %the differential equation for the evolution
the evolution equation of the moments of the spins, which in the case of the correlations gives %is
\begin{equation*}
    \frac{\der\!\av{s_i s_j}_{k}}{\der k} \!=\! -2 \av{s_i s_j}_{k}+ \av{s_i \text{tanh} (\bm J \bm s )_j}_{k}
    + \av{s_j \text{tanh} \left(\bm J \bm s \right)_i}_{k}.
\end{equation*}
The above equation is exact for any coupling matrix $\bJ$, but cannot be solved in general. In the MF regime, we can expand the $\tanh$ at first order and solve the set of linear differential equations. Since it is very similar to the Gaussian case, we only provide %write
the results. At leading order, the evolution upon sampling of the two-body correlations, in the eigenbasis of the couplings $\bm J$, can be written \begin{equation*}
    \frac{\der \av{x_\alpha x_\beta}_{k}}{\der k} \!=\! -\! \av{x_\alpha x_\beta}_k \left( 2-  J_\alpha - J_\beta \right) + 2 \delta_{\alpha \beta},
\end{equation*}
whose solution is
\begin{align*}
    \av{x_\alpha x_\beta}_k\! =\! 
         \frac{\delta_{\alpha \beta}}{1-J_\alpha} \!+\! \caja{\av{x_\alpha x_\beta}_0\!-\!\frac{\delta_{\alpha \beta}}{1-J_\alpha}}  e^{-(2- J_\alpha- J_\beta) \frac{k}{\tau}}.
\end{align*}
This analytical description is valid only in the high-temperature regime, where ${\rm max}\{J_\alpha\} < 1$. Nevertheless, in this regime we can reproduce the same analysis of the Gaussian model and reach the same conclusion: BMs trained with non-convergent MCMC  to compute the gradient fit models that reproduce the training set statistics, but only when sampled following the dynamics of the training: same initial conditions, transition rules, and number of steps $k$.

\section{Numerical experiments}
\begin{figure*}[t]
    \centering\includegraphics[height=5cm,trim=0 20 0 0]{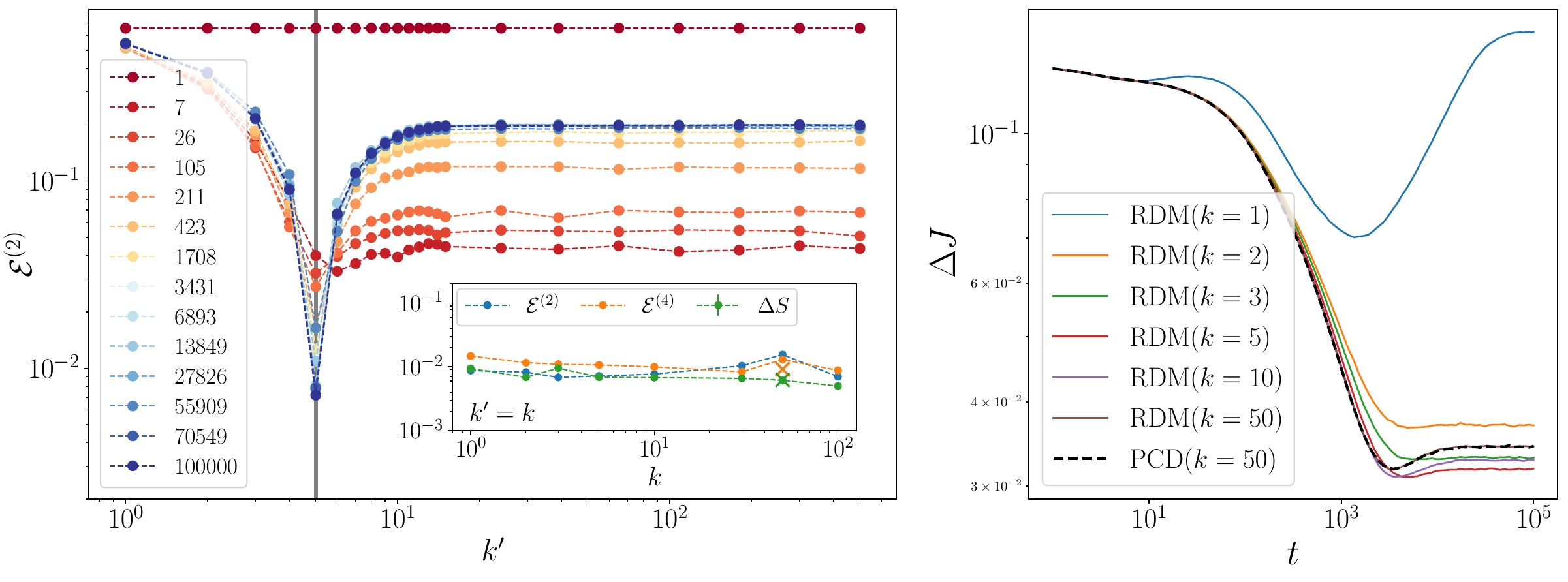}
    \caption{\textbf{Left:} Error over the covariance matrices $\mathcal{E}^{(2)}\!=\!\sum_{i<j}\textstyle(\av{x_i x_j}_{k^\prime,\bm p_0}\!-\!\av{x_i x_j}_{p_\mathcal{D}})^2 \slash \binom{N}{2}$ between the training set and the data generated at different learning ages vs the generation time $k'$, for a BM trained with data sampled from a $2$D ferromagnetic Ising model with $N\!=\!7^2$ at $k\!=\!5$. Similarly to Fig. \ref{fig:EvolJResampling} (right), the error shows a minimum at $k'\! =\! k$. \textbf{Inset:} Comparison of generated data's quality vs. the $k$ used for the learning. We show the error over covariance matrix, $\mathcal{E}^{(2)}$, the error over the $4-$th moments on $2-$sided plaquettes in the $2$D lattice, $\mathcal{E}^{(4)}$, and the relative entropy $\Delta S$ computed using gzip. Each of them is computed using samples generated in $k^\prime\!=\!k$ steps by machines trained for $t=10^5$ updates using different $k$. Cross-marks refer to the same result obtained by using persistent chains (PCD) with $k\!=\!50$. \textbf{Right:} Error over the learned couplings w.r.t. the true ones as a function of the training time $t$ for different values of $k$. A comparison with a BM trained using persistent chains (PCD) is also shown.
    The learning rate is set to $\gamma=10^{-2}$.
    \label{fig:BM_noneqeffect_2dFerro}}
\end{figure*}

This nonmonotonic effect in the quality of the generated samples and its relation to the mixing times of the algorithm was intensively studied numerically in the case of RBMs~\cite{decelle2021equilibrium} and yielded a picture fully compatible with the results obtained here. For example, the evolution of their ``quality estimators'' with sampling time is qualitatively very similar to those shown in Fig.~\ref{fig:EvolJResampling} for discrete sampling times. In Fig.~\ref{fig:rescifar}, we reproduce essentially the same results using the ConvNet from Ref.~\cite{Nijkamp_Hill_Han_Zhu_Wu_2020} (same architecture and hyperparameters) trained on the CIFAR-10~\cite{krizhevsky2009learning}. Here, we use the Frechet Inception Distance Score~\cite{fid} to evaluate the generation quality at different training times and as a function of sampling time.

Now we test our analytic predictions on the BM for the \textit{inverse Ising problem}, where $\bm x \in \{ -1,1 \}^N$, and the energy function is again quadratic in the variables $\bx$: $E_\bt (\bx)\! =\! -\!\sum_{i<j} J_{ij} x_i x_j\! -\! \sum_i h_i x_i$. In the following experiment, the dataset is made of configurations drawn from a $2$D ferromagnetic Ising model, with $N=7^2$ spins, $h_i = 0$, and $J_{ij} = \beta$ when $(i,j)$ are nearest neighbors on the $2$D lattice, zero otherwise. We consider $\beta = 0.44$, i.e a temperature slightly above the critical point. 
Learning is done with $k$-steps of nonconvergent heat-bath Markov chains initialized with random conditions. For $k=5$ and for machines obtained at different stages of the learning process, we show in Fig.~\ref{fig:BM_noneqeffect_2dFerro}--left the evolution of the error in the 2-point correlations for generated samples as a function of the number of MCMC steps used to synthesize them. We clearly observe the same behavior as in the Gaussian case, with a peak of best quality developing at $k' = k$.  We can also observe that the quality of the best generated samples does not depend on $k$, as shown in the inset of Fig.~\ref{fig:BM_noneqeffect_2dFerro}, where the error of $\mathcal{E}^{(2)}$ at $k'\!=\!k$ (at the end of the training) is plotted as a function of the $k$ used for training. Remarkably, we obtain the same result for two other observables that are not directly related to the generalized moments in the gradient, for example, the $4$-th order correlations between spins or the relative entropy (computed with gzip~\citep{Baronchelli_gzip}). Surprisingly,  trainings using $k\!=\!1$ generate samples as good as those with $k\!=\!50$. 
%As for the discussion of what happens along the training,
Concerning the behavior during training, we show in Fig.~\ref{fig:eigenvaluesBM} in the supplementary material the error in the first ten eigenvalues of the covariance matrix computed either with the real or the synthetic samples, just as shown for the Gaussian model in Fig.~\ref{fig:EvolJResampling}--right. Also in this case, the evolution of the error in the different eigenvalues in the early stages of learning has a minimum at $k^\prime\!\neq\! k$, as predicted, but the location of this peak does not necessarily coincide for the different eigenvalues. The early peaks are not as pronounced as in the Gaussian case because we do not have access to the best possible sampling time with discrete times, and the number of chains is relatively small. As training progresses, these best quality peaks gradually collapse towards $k^\prime\!=\! k$ for all eigenvalues.

Finally, in Fig.~\ref{fig:BM_noneqeffect_2dFerro}--right, we show the error in the inferred couplings $\bJ$ for different values of $k$ as a function of training time. These curves illustrate how very short times $k$ (e.g. $k\!=\!1$) can lead to very poor inferred models,
while 
models learned with short ($k\! \geq \!2$) nonconvergent chains can still catch important information about the data that is useful for interpretive applications.

\section{Conclusions}

Our results show that EBMs trained with non-persistent and non-convergent MCMCs to estimate the log-likelihood gradient are models with memory of their training scheme. We also show analytically how to exploit such memory effects to generate high-quality samples in a very short time, through a very precise sampling process that must perfectly match that of the training: i.e., same update rules, initializations and sampling time. EBMs trained with the contrastive divergence recipe fall into this category. However, they have the disadvantage that they cannot be used to generate new data because they provide minimal variation in the sample used for initialization. In this sense, we show that EBMs trained with short \emph{randomly} initialized MCMC runs can be effectively used as diffusion models for sample generation. \\On the other hand, we also prove that the equilibrium measure of an EBM trained in this way cannot correctly describe the statistics of the data and should be used with caution for interpretive applications. The lack of convergence of MCMC runs is particularly critical for datasets with an effectively low dimension, i.e., datasets that follow multimodal distributions, as mixing times become excessively long~\cite{bereux2022learning,decelle2021exact,nijkamp2022mcmc} to be thermalized by brute force.

\subsection*{Acknowledgements}
G.C, A.D. and B.S.  acknowledge financial support by the Comunidad de Madrid and the Complutense University of Madrid (UCM) through the Atracción de Talento programs (Refs. 2019-T1/TIC-13298 and 2019-T1/TIC-12776), the Banco Santander and the UCM (grant PR44/21-29937), and Ministerio de Econom\'{\i}a y Competitividad, Agencia Estatal de Investigaci\'on  and Fondo Europeo de Desarrollo Regional (Ref. PID2021-125506NA-I00). E.A. acknowledges support from the Swiss National Science Foundation by the SNSF Ambizione Grant No.~PZ00P2{\_}173962, and from the Simons Foundation (Grant No.~348126 to Sidney Nagel).
B.S. acknowledges the support of the Agence Nationale de la Recherche (Ref. ANR CPJ-ESE 2022).
\bibliography{icml_bib}
\bibliographystyle{icml2023}

%%%%%%%%%%%%%%%%%%%%%%%%%%%%%%%%%%%%%%%%%%%%%%%%%%%%%%%%%%%%%%%%%%%%%%%%%%%%%%%
%%%%%%%%%%%%%%%%%%%%%%%%%%%%%%%%%%%%%%%%%%%%%%%%%%%%%%%%%%%%%%%%%%%%%%%%%%%%%%%
% APPENDIX
%%%%%%%%%%%%%%%%%%%%%%%%%%%%%%%%%%%%%%%%%%%%%%%%%%%%%%%%%%%%%%%%%%%%%%%%%%%%%%%
%%%%%%%%%%%%%%%%%%%%%%%%%%%%%%%%%%%%%%%%%%%%%%%%%%%%%%%%%%%%%%%%%%%%%%%%%%%%%%%
\newpage
\appendix
\onecolumn

\section{Finite and discrete time sampling}\label{ap:discrete}

In Sec.~\ref{sec:master_eq} we discussed the non-convergent sampling of generic EBMs assuming time to be continuous. Here, for completeness, we provide the analogous derivation for discrete time $k$, i.e., \emph{number of Gibbs steps} during sampling.

For simplicity, we assume that the number of possible states of our model is finite, and that they are indexed by $a$. For finite and discrete time, the {\em master equation} states that the probability of being at $a$ at a time $k+1$ is as follows 
\begin{eqnarray}
  p_a^{(k+1)} = p_a^{(k)} + \sum_{b(\neq a)}w_{b \to  a}\, p_b^{(k)}
-\sum_{b(\neq a)}w_{a \to  b}\, p_a^{(k)} \label{eq:mastereq}
\end{eqnarray}
where $w_{a\to b}$ is the transition probability of going from state $a$ to $b$, and is model-dependent. These transition probabilities are independent of $k$ because the process is Markovian.
We introduce the vector notation ${\bm p}^{(k)}$, whose $a$-th component is $p_a^{(k)}$.
The evolution of the probabilities can be described by a {\em stochastic evolution} matrix $\U (\bt)$ with the following elements
\begin{eqnarray}
\U_{ab}=w_{b\to a}\ (a\neq b),
&\mathcal{U}_{aa}=1-\sum_{b\neq a}w_{a \to  b},
\end{eqnarray}
as
\begin{equation}
{\bm p}^{(k+1)}={ \bU}\,{\bm p}^{(k)}.
%\quad \Rightarrow \quad
%{\bm p}^{(k)}=\bU^k \,\bm p_{0}=\sum_\alpha c_\alpha\lambda_{\alpha}^k \bm u_{\alpha}.
\end{equation}
%where we have introduced a vector notation where $p_a^{(k)}$ is the $a$-th component of $\bm p^{(k)}$.
The probability distribution of a MCMC process after $k$ sampling steps can then be easily described in terms of the initial distribution ${\bm p}^{(0)}={\bm p}_0$ as 
\begin{equation}
\label{eq-discrete-time-proba-k}
{\bm p}^{(k)}=\bU^k \,\bm p_{0}=\sum_\alpha c_\alpha\lambda_{\alpha}^k \bm u_{\alpha},
={\bm u}_0 + \sum_{\alpha>0} c_\alpha\lambda_{\alpha}^k \bm u_{\alpha}
\end{equation}
where $\{ \lambda_{\alpha}\}$ and $\{{\bm u}_{\alpha}\}$ are the eigenvalues and eigenvectors of the $\bU$ matrix, and $ c_\alpha$ is the $\alpha$-th component of ${\bm p}^{(0)}$ in the same basis. Since the probability $\bm{p}^{(k)}$ must be normalized at all $k$ (i.e. $\sum_a p_a^{(k)}=1$), it is straightforward to check that the largest eigenvalue (namely, $\lambda_0$) must be 1, $\sum_a u_{0,a}=1$ and $c_0=1$, and that the rest must be $|\lambda_\alpha|<1$ for $\alpha\neq 0$ as long as {$\U$} 
is irreducible (i.e. any configuration can be reached from any other in a finite number of steps)\footnote{All elements of the $\mathcal{U}$ matrix are positive semi-definite, then, the Perron-Frobenius theorem guarantees that $\lambda_0=1$ is non degenerate if {$\mathcal{U}$} is irreducible.}. Then
\begin{equation}\label{eq:probevdisc}
 {\bm p}^{(k)}=\bm u_{0}+\sum_{\alpha>0}{\rm sign}{\lambda_\alpha} c_\alpha\, e^{-k|\log \lambda_\alpha|}\bm u_{\alpha},
\end{equation}
which means that it is guaranteed that the stochastic dynamics converge to the distribution $\bm u_0$ after a long time. Also, ${\kappa_\mathrm{mix}}=1/|\log \lambda_1|$ is now the {\em mixing} time, with $|\lambda_1|$ being the second largest eigenvalue. For the standard choice for the Gibbs sampling, the Heat Bath update, all eigenvalues  are nonnegative~\cite{dyer2014structure,guo2018layerwise}. 
As in the continuous case, we can impose $\bm u_0$ to be the model Gibbs distribution $\bm p_{\bt}$ by imposing detailed balance. This means that we can again write the time dependence of the probability in the form
\begin{equation}
\label{eq-pk-Lambda-discrete-time}
  {\bm p}^{(k)}=\bm p_{\bt}+\bm \Lambda_{\bt}^{(k)}(\boldsymbol{p}_0),
\end{equation}
where $\bm\Lambda_\bt$ is now the term $\sum_{\alpha > 0}$ in Eq.~\eqref{eq:probevdisc}.
From there, we can come back to Eq.~\eqref{eq:dynU} in the main text.

\section{Extension to continuous variables}\label{ap:contvar}

It is possible to extend the previous results to the case of continuous variables, in particular to analyze the evolution of the distribution of MCMC chains.
For simplicity, we will consider the discrete time case.

We consider continuous variables $\bx$ instead of the discrete set of variables denoted $a$ in the previous section.
Since the state space is continuous, the previous state probabilities $\lbrace p^{(k)}(a) \rbrace = {\bm p}^{(k)}$ are replaced by the probability density ${\bm p}^{(k)}(\bx)$ at time $k$.
%and the probabilities at time $k$ are replaced by the reversible Markov kernel $P(\bx,k)$.
The master equation can now be written as follows
\begin{eqnarray*}
    p^{(k+1)}(\bx) = \int d\bx' w(\bx|\bx') p^{(k)}(\bx')
\end{eqnarray*}
where $w(\bx|\bx')$ is the  kernel matrix for the transition from state $\bx'$ to state $\bx$. We now assume that the dynamics is irreducible and that the transition kernel satisfies the detailed balance conditions
\begin{equation*}
    p_{\rm eq}(\bx') w(\bx|\bx') = p_{\rm eq}(\bx) w(\bx'|\bx),
\end{equation*}
where $p_{\rm eq}(\bm x)$ will be  the Boltzmann distribution of the model, i.e, what we call $p_\bt(\bm x)$ in the main-text. 
In such a case, and this is where the subtleties of working with continuous versus discrete variables matters,
we can show that the operator $w$ is bounded and self-adjoint with respect to the inner product defined by the equilibrium distribution. These properties imply that the operator can be diagonalized and that the spectrum is real and bounded in $[-1,1]$, and the highest eigenvalue is $\lambda_{\rm max}=1$. We can show that the equilibrium distribution is the eigenvector associated to the highest eigenvalue $\lambda_{\rm max} = 1$ (as for the case of discrete states in the previous section):
\begin{equation*}
    \int d\bx' w(\bx | \bx') p_\bt(\bx') = \int d\bx' w(\bx' | \bx) p_\bt(\bx) = p_\bt(\bx).
\end{equation*}
Now, following the same line as in the article, we can decompose any initial probability distribution in the complete basis of continuous eigenfunctions $\lbrace \phi_{\alpha}(\bx)\rbrace_\alpha$
as
\begin{eqnarray*}
    p_0(\bx) = \sum_{\alpha=0}^{\infty} c_\alpha \phi_\alpha(\bx) = p_\bt(\bx) + \sum_{\alpha=1}^{\infty} c_\alpha \phi_\alpha(\bx).
\end{eqnarray*}
From this point, by using
the repeated application of the master equation
to relate the distribution at time $k$ to the initial distribution, we obtain as in the previous section at Eq.~\eqref{eq-discrete-time-proba-k}:
\begin{equation*}
    p^{(k)}(\bx) = p_\bt(\bx) + \sum_{\alpha=1}^{\infty} c_\alpha \lambda_\alpha^k \phi_\alpha(\bx).
\end{equation*}
From this point on, the same derivations apply as in the main text, but as in Eq.~\eqref{eq-pk-Lambda-discrete-time} the definition of $\bm \Lambda_{\bt}$ is now
\begin{equation}
    \bm \Lambda_{\bt}^{(k)}(\bm x;\boldsymbol{p}_0)=\sum_{\alpha=1}^{\infty} c_\alpha \lambda_\alpha^k \phi_\alpha(\bx).
\end{equation}

\section{Proof of theorem \ref{thm:conv_gauss}} \label{ap:thm_k}

We want to analyze the solution of Eq. \eqref{eq:fixedpoint_gauss}.
{In order to do that,} %To to that,
we define the following function
\begin{equation}
    g(J) = J \hat{c}_{\alpha \alpha} - 1 + e^{- 2 J k} (1 - J \av{x_\alpha^2}_0 ) .
\end{equation}
We can compute the first and second order derivative
\begin{align}
    g'(J) &= \hat{c}_{\alpha \alpha} + e^{- 2 J k}\left( 2Jk \av{x_\alpha^2}_0 - 2k - \av{x_\alpha^2}_0 \right), \\
    g''(J) &= e^{- 2 J k} \left( 4k \av{x_\alpha^2}_0 + 4k^2  - 4Jk^2 \av{x_\alpha^2}_0 \right).
\end{align}
The second order derivative cancels at
\begin{equation}
    J_0 = \frac{ k\av{x_\alpha^2}_0 +1}{k  \av{x_\alpha^2}_0 }.
\end{equation}
Therefore for large negative value of $J$, $g''$ is positive and decreasing. It becomes negative for $J>J_0$ and then converge towards zero, {remaining} negative. Thus, $g'$ is an increasing function up to $J_0$, and then it decreases towards $\hat{c}_{\alpha \alpha}$. We are interested in knowing where the first order derivative crosses $J=0$. It can be for negative values of $J$, in that case the dynamics converges to a negative value $J^*$, or for positive values of $J$, in that case the dynamics converges to a positive value $J^*$. Given the variation of the function, it is enough to look at the sign of the first derivative at $J=0$.
\begin{equation}
    g'(0) = 0 \Longleftrightarrow k^* = \frac{\hat{c}_{\alpha \alpha} - \av{x_\alpha^2}_0}{2}.   
\end{equation}

\section{Dynamical equations with rotations} \label{ap:rot}

We consider a simplified case in two dimensions. We consider the eigenspace of the covariance matrix given by two orthogonal vectors $\hat{\bu}_{1,2}$, and the ones of the weight matrix $\bu_{1,2}$, and we denote by $\phi (t)$, the angle between the vectors $\hat{\bu}_{1}$, $\bu_{1}$ ($\hat{\bu}_{2}$, $\bu_{2}$.respectively)
\begin{align}
    \bu_1 &= \hat{\bu}_1 \cos(\phi) + \hat{\bu}_2 \sin(\phi) \\
    \bu_2 &= -\hat{\bu}_1 \sin(\phi) + \hat{\bu}_2 \cos(\phi) 
\end{align}
When the vectors are not aligned, and using the eigendecomposition of the covariance matrix: $\av{x_i x_j}_\pemp = \sum_\gamma \hat{u}_i^\gamma \hat{c}_\gamma \hat{u}_j^\gamma$, we have that
\begin{equation}
    \hat{c}_{\alpha \beta} = \sum_{i,j} u_i^\alpha u_j^\beta \av{x_i x_j}_\pemp = \sum_{i,j,\gamma} u_i^\alpha u_j^\beta \hat{u}_i^\gamma \hat{c}_\gamma \hat{u}_j^\gamma
\end{equation}
Therefore we have
\begin{align}
    \hat{c}_{11} &= \hat{c}_1 \cos(\phi)^2 + \hat{c}_2 \sin(\phi)^2 \\
    \hat{c}_{22} &= \hat{c}_1 \sin(\phi)^2 + \hat{c}_2 \cos(\phi)^2 \\
    \hat{c}_{12} &= -\hat{c}_1 \cos(\phi) \sin(\phi) + \hat{c}_2 \cos(\phi) \sin(\phi) = \cos(\phi) \sin(\phi) (\hat{c}_2 - \hat{c}_1)
\end{align}
Now, given the parametrization of $\bu_{1,2}$, we can see that
\begin{equation}
    \bm{u}_1 \frac{d\bu_2}{dt} = -\frac{d \phi}{dt}
\end{equation}
Using Eq. \eqref{eq:grad_Ja},\eqref{eq:grad_ua},\eqref{eq:corrLangeving}, we obtain the dynamical equations of this simplified system.

\section{Supplementary Figures}

\begin{figure}[!h]
    \centering\includegraphics[width=\textwidth]{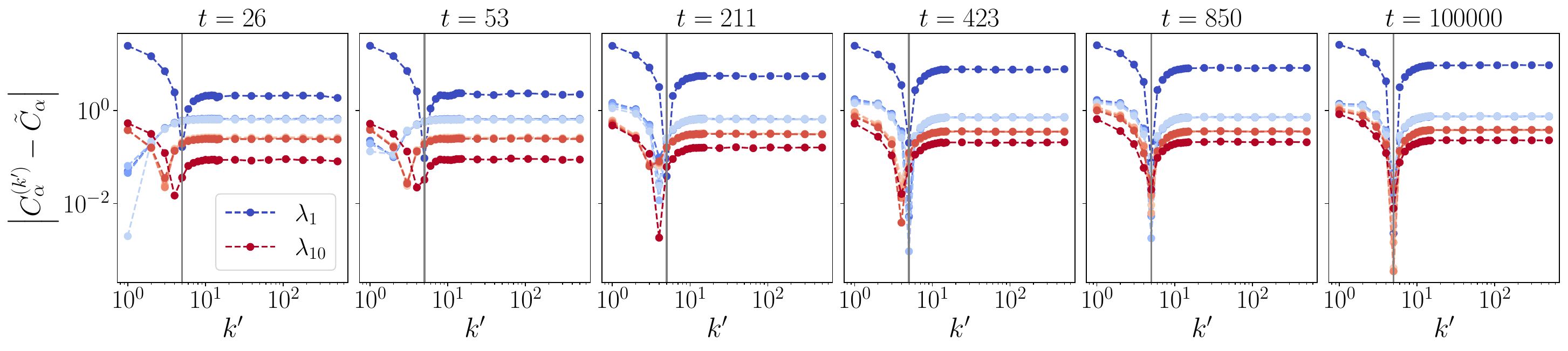}
    \caption{Error over the eigenvalues of the covariance matrix generated after $k'$ steps of MCMC vs the dataset covariance matrix. The setting is the same of Figure \ref{fig:BM_noneqeffect_2dFerro} in the main text. Each panel shows the element-wise error between the eigenvalues of $\boldsymbol{C}^{(k')}$ and $\tilde{\boldsymbol{C}}$ (at different stages of the learning $t$ shown at the top of each panel), computed for the first $10$ largest (in absolute value) eigenvalues. Some eigenvalues are degenerated and superposed.
    \label{fig:eigenvaluesBM}}
\end{figure}

\end{document}